\documentclass[letterpaper, 10 pt, journal, twoside]{IEEEtran}
\usepackage{amsmath,amsfonts}
\usepackage{algorithmic}
\usepackage{algorithm}
\usepackage{array}
\usepackage[caption=false,font=normalsize,labelfont=sf,textfont=sf]{subfig}
\usepackage{textcomp}
\usepackage{stfloats}
\usepackage{url}
\usepackage{verbatim}
\usepackage{graphicx}
\usepackage{cite}
\usepackage{xcolor}
\usepackage{multirow}
\usepackage{makecell}    
\usepackage{dsfont}  
\usepackage{float} 
\usepackage{amssymb}
\usepackage{booktabs}
\usepackage{wrapfig}
\usepackage{lipsum} 
\usepackage{subcaption} 
\usepackage{hyperref} 

\begin{document}

\title{FR-Net: Learning Robust Quadrupedal Fall Recovery on Challenging Terrains through Mass-Contact Prediction}
\author{Yidan Lu$^{*}$, Yinzhao Dong$^{*}$, Jiahui Zhang, Ji Ma, Peng Lu$^{\dagger}$%
\thanks{Manuscript received: December 31, 2024; Revised: March 27, 2025; Accepted April 24, 2025.}
\thanks{This paper was recommended for publication by Editor Jens Kober upon evaluation of the Associate Editor and Reviewers' comments.
This work was supported by the General Research Fund under Grant 17204222, and in part by the Seed Fund for Collaborative Research and General Funding Scheme--{HKU-TCL} Joint Research Center for Artificial Intelligence.} 
\thanks{The authors are with the Adaptive Robotic Controls Lab (ArcLab), Department of Mechanical Engineering, The University of Hong Kong, Hong Kong SAR, China. \protect\url{ydlu@connect.hku.hk}, \protect\url{dongyz@connect.hku.hk}, \protect\url{holmesz@connect.hku.hk}, \protect\url{maji@connect.hku.hk}.}
\thanks{$^{*}$ Equal Contribution. $^{\dagger}$Corresponding author: \protect\url{lupeng@hku.hk,}} 
\thanks{The supplementary video is available at \protect\url{https://youtu.be/rG1Zf7IwJdc}}
\thanks{Digital Object Identifier (DOI): 10.1109/LRA.2025.3569117}
}

\markboth{IEEE Robotics and Automation Letters. Preprint Version. Accepted April, 2025}
{Lu \MakeLowercase{\textit{et al.}}: FR-Net: Learning Robust Quadrupedal Fall Recovery on Challenging Terrains through Mass-Contact Prediction} 


\maketitle

\begin{abstract}
Fall recovery for legged robots remains challenging, particularly on complex terrains where traditional controllers fail due to incomplete terrain perception and uncertain interactions. We present \textbf{FR-Net}, a learning-based framework that enables quadrupedal robots to recover from arbitrary fall poses across diverse environments. Central to our approach is a Mass-Contact Predictor network that estimates the robot's mass distribution and contact states from limited sensory inputs, facilitating effective recovery strategies. Our carefully designed reward functions ensure safe recovery even on steep stairs without dangerous rolling motions common to existing methods. Trained entirely in simulation using privileged learning, our framework guides policy learning without requiring explicit terrain data during deployment. We demonstrate the generalization capabilities of \textbf{FR-Net} across different quadrupedal platforms in simulation and validate its performance through extensive real-world experiments on the Go2 robot in 10 challenging scenarios. Our results indicate that explicit mass-contact prediction is key to robust fall recovery, offering a promising direction for generalizable quadrupedal skills.
\end{abstract}

\begin{IEEEkeywords}
Reinforcement learning, legged robots, failure detection and recovery.
\end{IEEEkeywords}

\section{Introduction}
\begin{figure}[t]
   \centering
   \includegraphics[width=0.47\textwidth, height=0.73\textwidth]{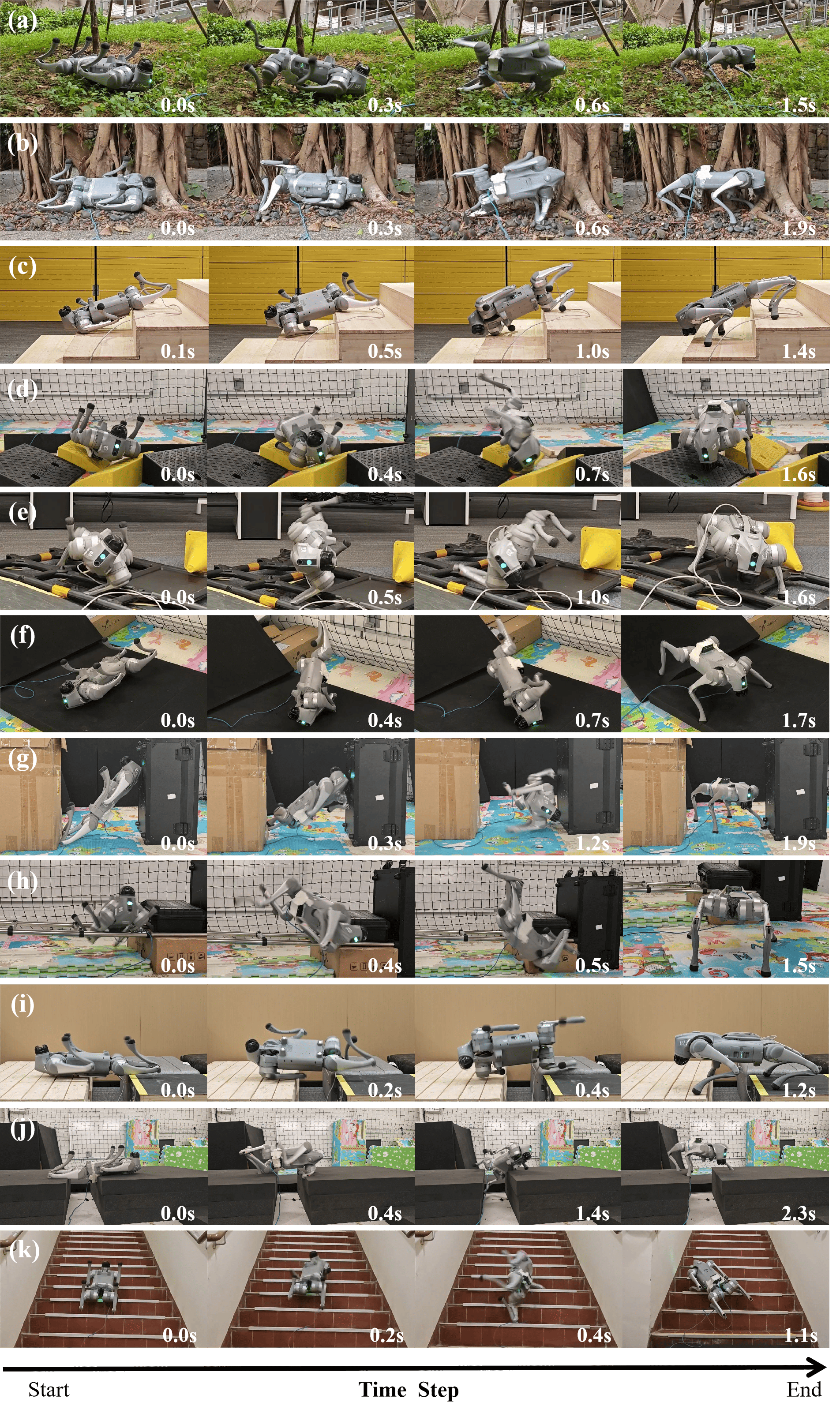}
   \caption{\textcolor{black}{Go2 quadrupedal robot recovering from prone positions across various terrains: (a) Grass field, (b) Small stones, (c) Simple stairs, (d) Messy floor, (e) Uneven terrain, (f) Foam cushion, (g) Discrete obstacles, (h) Air beams, (i) Solid gaps, (j) Risky gaps, and (k) Stiff stairs. Columns show recovery phases over time.}}
   \label{fig:real_results}
\end{figure}

\IEEEPARstart{T}{he} diversity and unpredictability of complex terrains, such as steep slopes, cantilever beams, and gaps, pose significant challenges to the locomotion of legged robots \cite{abdalla2023efficient, zhang2023learning, luo2024moral, cheng2024extreme,hoeller2024anymal, luo2024pie}. These environments can cause falls due to intricate and uneven surfaces that alter balance and stability. Therefore, the ability of a robot to autonomously recover from falls is of paramount practical significance. It not only enhances the efficiency of task completion by reducing downtime but also ensures the safety of the robot by preventing further slips or tumbles that could lead to damage. In mission-critical applications such as search and rescue or exploration in hazardous environments, the capability to recover autonomously from falls becomes even more crucial. This ability allows robots to continue their tasks without human intervention, saving valuable time and resources. In dangerous areas, this capability can prevent further accidents, thereby safeguarding both the robot and the mission success \cite{hwangbo2019learning, arm2023scientific, jenelten2024dtc}. The challenges are further complicated by the diverse nature of fall scenarios, from simple tumbles to complex situations involving multiple contact points, each requiring different recovery strategies.

\IEEEpubidadjcol

Current methodologies addressing autonomous fall recovery can be broadly categorized into model-based and learning-based approaches. Currently, model-based approaches predominantly rely on simplified models and predefined contact points, rendering them ineffective in managing non-smooth dynamic contact scenarios \cite{khorram2015balance}, \cite{focchi2018slip}. This limitation restricts the ability of robots to recover autonomously from falls in complex terrains. In contrast, learning-based methods, particularly those employing deep reinforcement learning (DRL), have demonstrated strong performance in recovery tasks on flat or moderately uneven terrains. However, these approaches still exhibit shortcomings in developing effective recovery strategies for highly irregular and unpredictable terrains \cite{lee2019robust, zhang2023research, nahrendra2023robust}. These developments underscore the potential of DRL-based approaches to overcome the inherent limitations of traditional model-based methods, offering more flexible and resilient solutions for autonomous fall recovery in a wide array of unpredictable and complex environments.

\begin{figure*}[h]
\centering\includegraphics[width=1.0\textwidth,height=0.33\textwidth]{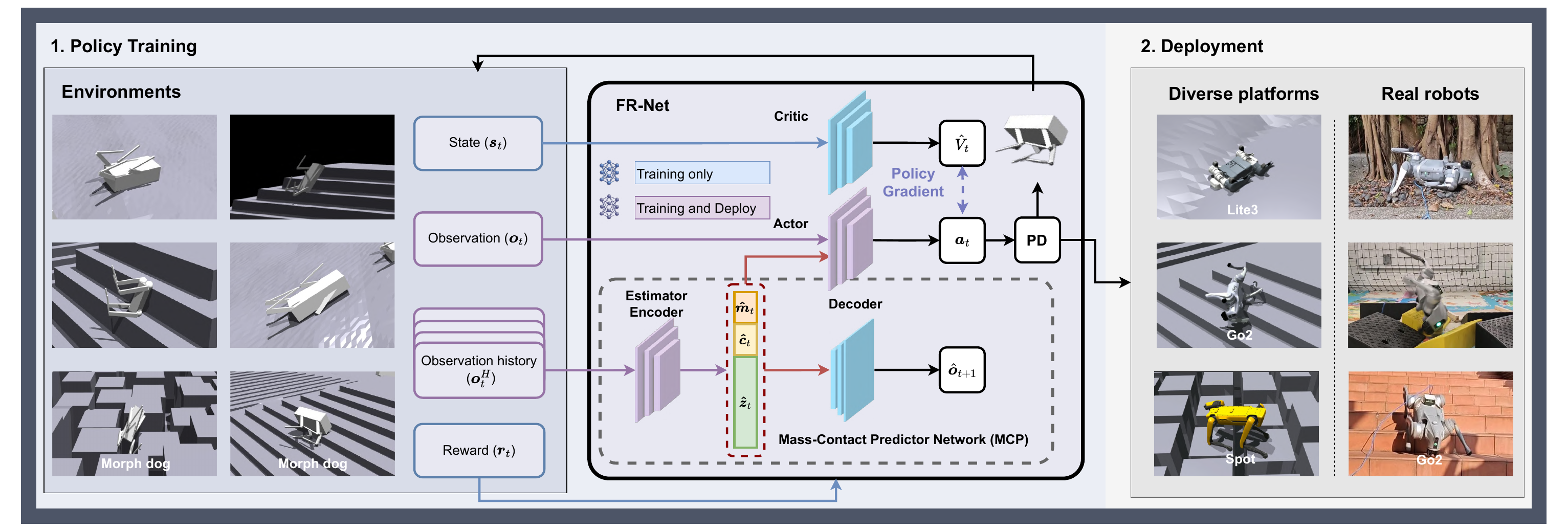}
\caption{Overview of the FR-Net framework. Trained on morphological dogs could be deployed on multiple platforms.}
\label{pics_framework}
\end{figure*}

In this paper, we propose \textbf{F}all \textbf{R}ecovery \textbf{Net}work (\textbf{FR-Net}), a learning-based framework that integrates a mass-contact predictor network with meticulously designed reward functions. FR-Net aims to enhance the fall recovery capabilities of quadrupedal robots in challenging terrains. The key contributions of this work are threefold:

\begin{itemize}
    \item A novel Mass-Contact Predictor network architecture that effectively estimates robot state from limited sensory inputs, enabling robust recovery strategy generation across diverse terrains.
    
    \item A carefully designed reward function that, combined with our framework, ensures safe recovery motions even on steep terrains without dangerous rolling behaviors that plague existing approaches.
    
    \item \textcolor{black}{Comprehensive validation of our framework's generalization capabilities through cross-platform testing in simulation (Go2, Spot, Lite3, et al.) and extensive real-world experiments on the Go2 robot, demonstrating superior recovery performance across challenging scenarios.}
\end{itemize}

\section{Related Work}

\subsection{\textcolor{black}{Model-Based Fall Recovery for Quadruped Robots}}
Model-based methods rely on dynamic models and control algorithms to design recovery strategies. For instance, Yang et al.~\cite{yang2018falling} proposed a fall prediction model to identify potential falls during locomotion, allowing for proactive recovery actions. Khorram et al.~\cite{khorram2015balance} developed balance recovery strategies using simplified models of bipedal robots. While these approaches can be effective in controlled settings, they often depend on simplified models and predefined contact points, limiting their adaptability to complex and unpredictable terrains. Additionally, traditional optimal control algorithms may struggle with non-smooth dynamics and unexpected contact scenarios that occur during falls.
\subsection{\textcolor{black}{Learning-Based Fall Recovery for Legged Robots}}
Learning-based methods, particularly those utilizing deep reinforcement learning (DRL), have shown promise in handling the complexities of fall recovery without explicit modeling of collision dynamics \cite{zhang2023research}, \cite{li2024dynamic}, \cite{yang2023learning}, \cite{gaspard2024frasa}. Hwangbo et al.~\cite{hwangbo2019learning} demonstrated that DRL can be used to learn dynamic motor skills for quadrupedal robots, including recovery maneuvers on flat terrains. \textcolor{black}{Lee et al.~\cite{lee2019robust} further developed robust recovery controllers for quadrupedal robots using deep reinforcement learning, enabling effective getting-up behaviors from arbitrary fallen configurations. Smith et al.~\cite{smith2022legged} developed a recovery controller serving as a reset mechanism for locomotion policies, but it remains primarily effective on flat surfaces. Nahrendra et al. \cite{nahrendra2023robust} proposed DreamRiser, a robust all-terrain recovery motion control framework for quadrupedal robots that leverages learned terrain imagination.} More recent efforts, such as the DribbleBot system~\cite{ji2023dribblebot}, showcased recovery capabilities on various outdoor surfaces like sand and gravel. \textcolor{black}{Ma et al.~\cite{ma2023learning} proposed an arm-assisted framework for fall damage reduction and recovery in legged mobile manipulators. Scalise et al.~\cite{scalise2024toward} developed a Student-Teacher RL approach for quadruped self-righting on flat, rough, and inclined surfaces.}


While multi-sensor systems incorporating cameras and LiDARs can enhance environmental awareness~\cite{miki2022learning, margolis2024rapid}, they require additional hardware that may be vulnerable during falls. Alternative approaches focus on leveraging proprioceptive information and interaction forces to estimate terrain properties~\cite{fu2022coupling}, \textcolor{black}{but they often struggle to accurately perceive and interpret complex environmental features.} Despite these advancements, a significant gap remains in developing robust recovery strategies for highly challenging terrains, such as steep stairs, cantilever beams, and gaps. Existing methods may not adequately handle the diverse physical properties and complex geometries encountered in these scenarios, highlighting the need for approaches that can adapt to a wide range of environmental conditions while maintaining reliable perception capabilities.

\section{Quadrupedal Fall Recovery Controller}

As shown in Fig. \ref{pics_framework}, we propose FR-Net that enables quadrupedal robots to perform effective fall recovery in complex scenarios. The FR-Net consists of three sub-networks: Mass-Contact Predictor Network, Actor Network, and Critic Network. These components work together to achieve real-time adaptation to various terrains and robot dynamics. Next, each part of the controller will be discussed in detail.

\subsection{Asymmetric Actor-Critic Network}
Certain privileged information is vital for fall recovery tasks. For instance, on highly sloped terrains, the position of the robot's center of mass relative to its support polygon is essential for maintaining stability, as even the slightest deviations can lead to uncontrollable rolling. In discontinuous gaps, accurate terrain perception is vital for safe foot placement and preventing potential falls into voids. Thus, the fall recovery problem of the quadrupedal robot can be formulated as a Partially Observable Markov Decision Process (POMDP) \cite{spaan2012partially}. Accordingly, our controller adopts the asymmetric actor-critic structure for terrain-aware robot learning, enabling more effective exploration under high-dimensional spaces of legged robots. The actor and critic learn from distinct objectives: the former aims to maximize expected rewards, while the latter minimizes the difference between predicted and actual values.

\subsubsection{Actor network}
The actor network $\pi_\theta (\boldsymbol{a}_t|\boldsymbol{p}_t)$ is parameterized by $\theta$ and maps the augmented state  $\boldsymbol{p}_t \in\mathbb{R}^{75}$ to control actions $\boldsymbol{a}_t\in\mathbb{R}^{12}$, i.e.
\begin{equation}
\boldsymbol{p}_t = 
 \begin{bmatrix}
 \boldsymbol{o}_t & \boldsymbol{\hat{m}}_t & \boldsymbol{\hat{c}}_t &
 \boldsymbol{\hat{z}}_t
 \end{bmatrix}^T
\end{equation}
where the proprioceptive observation $\boldsymbol{o}_t \in\mathbb{R}^{42}$ consists of body angular velocity $\boldsymbol{\omega}_t\in\mathbb{R}^{3}$, projected gravity vector $\boldsymbol{g}_t\in\mathbb{R}^{3}$, joint angles $\boldsymbol{q}_t\in\mathbb{R}^{12}$, joint velocities $\dot{\boldsymbol{q}}_t\in\mathbb{R}^{12}$, and the previous action $\boldsymbol{a}_{t-1}\in\mathbb{R}^{12}$, all collected from onboard sensors. The Mass-Contact Predictor Network outputs $\boldsymbol{\hat{m}}_t \in \mathbb{R}^{4}$ representing the predicted mass distribution across base, hip, thigh and calf, $\boldsymbol{\hat{c}}_t \in \mathbb{R}^{13}$ indicating the collision probabilities of body components, and $\boldsymbol{\hat{z}}_t \in \mathbb{R}^{16}$ encoding a latent representation.

The action $\boldsymbol{a}_t\in\mathbb{R}^{12}$ represents the desired increment of the joint angle w.r.t the initial pose $\boldsymbol{\mathring{q}}$, i.e. $\boldsymbol{q}^{*}_{t} = \boldsymbol{\mathring{q}} + \boldsymbol{a}_{t}$. The final desired angle $\boldsymbol{q}^{*}_{t}$ is tracked by the torque generated by the joint-level PD controller of the joint-level actuation module in the simulator, i.e. $\boldsymbol{\tau} = \boldsymbol{k}_p \cdot (\boldsymbol{q}^{*}_{t} - \boldsymbol{q}_t) + \boldsymbol{k}_d \cdot ( - \boldsymbol{\dot{q}}_{t})$.

\subsubsection{Critic network}

The critic network $V_\psi(\boldsymbol{s}_t)$, parameterized by $\psi$, estimates the state value function by taking privileged observation $\boldsymbol{s}_t$ as input, which is defined as:
\begin{equation}
\mathbf{s}_t=
\begin{bmatrix}
\mathbf{o}_t & \mathbf{h}_t & \mathbf{m}_t & \boldsymbol{k}_\text{PD} & \boldsymbol{p_{com}} & \boldsymbol{c}_t & \boldsymbol{c_f} & \boldsymbol{\mu} 
\end{bmatrix}^T
\end{equation}
where $\boldsymbol{o}_t \in \mathbb{R}^{42}$ denotes proprioceptive observation, $\boldsymbol{h}_t \in \mathbb{R}^{187}$ represents the height map scan dots, $\boldsymbol{m}_t \in \mathbb{R}^4$ contains the masses of the base, hip, thigh, and calf, $\boldsymbol{k}_\text{PD} \in \mathbb{R}^{24}$ stores the Proportional-Derivative (PD) control gains, $\boldsymbol{p}_\text{com} \in \mathbb{R}^2$ represents the center of body mass, $\boldsymbol{c}_t \in \mathbb{R}^{13}$ indicates contact states of each component, $\boldsymbol{c}_f \in \mathbb{R}^4$ denotes contact forces of each foot, and $\boldsymbol{\mu} \in \mathbb{R}^1$ is the friction coefficient.

\subsection{Mass-Contact Predictor Network}

\textcolor{black}{Beyond privileged information, explicit terrain perception is crucial for legged locomotion. While conventional controllers rely on dedicated sensors like cameras \cite{luo2024pie} or LiDAR \cite{wisth2022vilens}, Cheng et al. \cite{cheng2024quadruped} utilized the possibility of inferring terrain information through contact interactions. Additionally, Luo et al. \cite{luo2024moral} demonstrated that real-time mass distribution estimation can enhance controller adaptability across robots with different morphologies. Inspired by these works, we propose a \textbf{M}ass-\textbf{C}ontact \textbf{P}redictor Network (\textbf{MCP}) which leverages proprioceptive information to understand the surrounding terrain while simultaneously estimating the mass distribution across different robot components.}

\subsubsection{Predictor Network}
The MCP consists of an Encoder and a Decoder. The Encoder $E_\phi(\boldsymbol{o}_t^H)$, parameterized by $\phi$, processes $H$ consecutive observation frames to estimate the robot's mass distribution, the contact probabilities of its body components, and to extract temporal features:

\begin{equation}
\begin{bmatrix}
\boldsymbol{\hat{m}}_t & \boldsymbol{\hat{c}}_t & \boldsymbol{\hat{z}}_t  \end{bmatrix}^T = E_\phi(\boldsymbol{o}_t^H)
\end{equation}
where $\boldsymbol{o}_t^H = \begin{bmatrix} \boldsymbol{o}_{t-H+1} \ldots \boldsymbol{o}_{t-1} & \boldsymbol{o}_{t} \end{bmatrix}^T$ is a temporal observation sequence with $H=5$ frames in this task. The Decoder $D_\theta$ utilizes the compressed representations from the encoder to predict the subsequent observation. Specifically, it use the estimated masses $\boldsymbol{\hat{m}}_t$, estimated collision probabilities $\boldsymbol{\hat{c}}_t$, and the latent vector $\boldsymbol{\hat{z}}_t$ to generate the predicted next frame observation $\boldsymbol{\hat{o}}_{t+1}$:
\begin{equation}
\boldsymbol{\hat{o}}_{t+1} = D_\theta(\boldsymbol{\hat{m}}_t, \boldsymbol{\hat{c}}_t,  \boldsymbol{\hat{z}}_t)
\end{equation}
Here, $\boldsymbol{\hat{o}}_{t+1} \in \mathbb{R}^{42}$ is the predicted observation at time $t+1$, and $\boldsymbol{o}_{t+1} \in \mathbb{R}^{42}$ represents the ground truth observation.

\subsubsection{Loss Function}
The training objective comprises multiple loss terms that ensure the accuracy of the estimations and the meaningfulness of the latent representations. The total loss $\mathcal{L}_{\varphi}$ is partitioned into Estimation Loss and Variational Autoencoder (VAE) Loss.
\begin{equation}
\mathcal{L}_{\varphi} = \mathcal{L}_{\varphi}^{est} + \mathcal{L}_{\varphi}^{VAE}
\end{equation}

The Estimation Loss $\mathcal{L}_{\varphi}^{est}$ includes mass estimation loss and collision estimation loss. A regression loss minimizes the discrepancy between the estimated masses and the ground truth. A Binary Cross-Entropy (BCE) loss reduces the difference between the actual contact information and the estimated collision state:
\begin{equation}
\begin{aligned}
\mathcal{L}_{\varphi}^{est} &= \lambda_{m}\, \text{MSE}(\boldsymbol{\hat{m}}_t, \boldsymbol{m}) \\
&\quad + \lambda_{c}\left( \boldsymbol{c}_t \log \boldsymbol{\hat{c}}_t + (1 - \boldsymbol{c}_t) \log (1 - \boldsymbol{\hat{c}}_t) \right)
\end{aligned}
\end{equation}
where $\lambda_{m}$ and $\lambda_{c}$ are hyperparameters that balance the contributions of the respective loss terms.

The VAE framework introduces two additional loss terms: Reconstruction Loss, which measures the accuracy of the predicted observations, and Kullback-Leibler (KL) Divergence Loss, which regularizes the latent space distribution to follow a standard normal distribution.
\begin{equation}
\begin{aligned}
\mathcal{L}_{\mathrm{VAE}} &= \lambda_{\text{rec}}\, \text{MSE}(\hat{\mathbf{o}}_{t+1}, \mathbf{o}_{t+1}) \\
&\quad + \lambda_{\text{KL}}\, D_{\mathrm{KL}}(q(\mathbf{z}_{t}|\mathbf{o}_{t}^{H}) \parallel p(\mathbf{z}_{t}))
\end{aligned}
\end{equation}
where $q(\mathbf{z}_t|\mathbf{o}_t^H)$ represents the posterior distribution of $\mathbf{z}_t$ given $\mathbf{o}_{t}^{H}$, and $p(\mathbf{z}_t)$ is the prior distribution, parameterized by a Gaussian distribution. A standard normal distribution is used for the prior distribution, and all observations are normalized to have zero mean and unit variance. $\lambda_{\text{rec}}$ and $\lambda_{\text{KL}}$ are hyperparameters that balance the respective loss terms.

\subsection{Reward Function Design}
The reward functions are designed to encourage robots to learn to stand in extremely complex terrain while maintaining safety, which are structured into four main categories as shown in Table \ref{tab:reward_terms}. Next, the impact of each category on fall recovery will be discussed in detail.

\begin{figure}[h]
    \centering
    \includegraphics[width=0.5\textwidth, height=0.1\textheight]{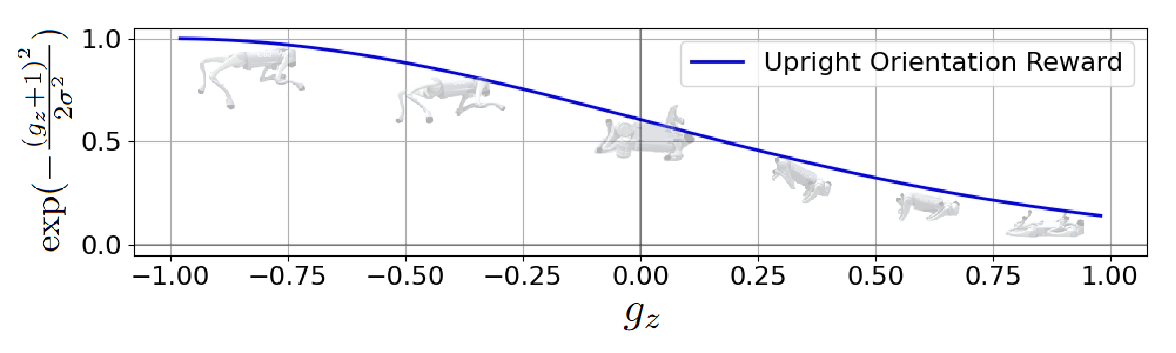}
    \caption{Upright Orientation Reward function and corresponding robot pose. The reward is computed as $\exp(-\frac{(g_z + 1)^2}{2\sigma^2})$, reaching its maximum at the upright position ($g_z \approx -0.98$).}
    \label{fig:ggseq}
\end{figure}

\subsubsection{Orientation Posture}
This category integrates multiple orientation-related rewards to maintain a stable posture and guide upright recovery behavior. The base orientation component ($\boldsymbol{g}_{xy}^2$) promotes stable body posture by penalizing excessive tilt from the horizontal plane, which is crucial for maintaining balance during the recovery process. The upright orientation reward, implemented through a Gaussian-based function $\exp(-\frac{(g_z + 1)^2}{2\sigma^2})$, facilitates recovery behavior by creating a gradient that approaches zero when the robot's ventral surface faces upward ($g_z \approx 0.98$), and reaches maximum when the robot is upright ($g_z \approx -0.98$), as shown by the blue curve in Fig. \ref{fig:ggseq}. Additionally, when the robot achieves near-upright orientation (defined by threshold $\epsilon$), a postural convergence term $\exp(-|\boldsymbol{q} - \boldsymbol{q}_{stand}|^2)$ encourages convergence to a predetermined reference posture. These components are combined with appropriate weights:

\begin{equation}
\begin{aligned}
R_{orientation} = \omega_1\boldsymbol{g}_{xy}^2 + \omega_2\exp(-\frac{(g_z + 1)^2}{2\epsilon^2}) \\
+\omega_3\exp(-(\boldsymbol{q} - \boldsymbol{q}_{stand})^2) \cdot \mathds{1}\{{|g_z + 1| < \epsilon}\}
\end{aligned}
\end{equation}

where $\omega_1$, $\omega_2$, and $\omega_3$ are weighting coefficients that balance the influence of each component in the overall orientation control strategy.
\subsubsection{Contact Management}
The contact management component focuses on maintaining appropriate contact states between the robot and the environment through two key mechanisms. The feet contact term $\sum_{i=1}^4 \mathds{1}_{contact_i}$ introduces a discrete reward that sums the binary ground contact indicators across all four feet, promoting stable recovery through appropriate foot placement. The body contact term $\mathds{1}_{contact}$ discourages any undesired contact between the robot's body (excluding feet) and the environment, helping prevent potentially dangerous collisions or unstable configurations. These components are combined as:
\begin{equation}
R_{contact} = \omega_4 \cdot  \sum_{i=1}^4 \mathds{1}_{contact_i} + \omega_5 \cdot \mathds{1}_{contact}
\end{equation}
where $\omega_4$ and $\omega_5$ are weighting factors that balance the importance of maintaining proper foot contact while avoiding unwanted body contact.

\subsubsection{Stability Reward}
The stability control terms ensure safe and controlled recovery behavior through force regulation and position maintenance. \textcolor{black}{Similar to stability mechanisms in \cite{scalise2024toward}, we incorporate terms to promote recovery on complex terrains. The safety force regulation term $\sum_{i=1}^{4} ||\boldsymbol{f}_{i}^{xy}||_2$ penalizes horizontal forces at knee contacts, where $\boldsymbol{f}_{i}^{xy}$ represents the horizontal components of contact forces at the $i$-th knee joint. This helps improve recovery behavior on stairs and slopes, where horizontal forces against vertical surfaces can generate adverse rolling moments.} The body position bias is implemented through a clipped Euclidean distance reward:
\begin{equation}
\begin{aligned}
R_{stability} = & \omega_6 \cdot \sum_{i=1}^{4} ||\boldsymbol{f}_{i}^{xy}||_2 \\ & + \omega_7 \cdot \text{clip}(||\boldsymbol{p}_{current}^{xy} - \boldsymbol{p}_{init}^{xy}||_2, 0, 4)\label{eq:R_stability}
\end{aligned}
\end{equation}

where $\boldsymbol{p}_{current}^{xy}$ and $\boldsymbol{p}_{init}^{xy}$ represent the current and initial horizontal positions of the robot's main body, and $\omega_6$, $\omega_7$ are the corresponding weights. \textcolor{black}{The force regulation prevents unsafe interactions with the environment, while the body-bias penalty mechanism discourages undesired displacement during recovery attempts. Unlike \cite{scalise2024toward}, our method explicitly considers additional constraints to adapt to steeper stairs and discontinuous gaps, ensuring recovery behavior remains consistent under extreme conditions.}

\begin{figure}[h]
\centering
\begin{minipage}[t]{0.48\columnwidth}
\centering
\includegraphics[width=\textwidth]{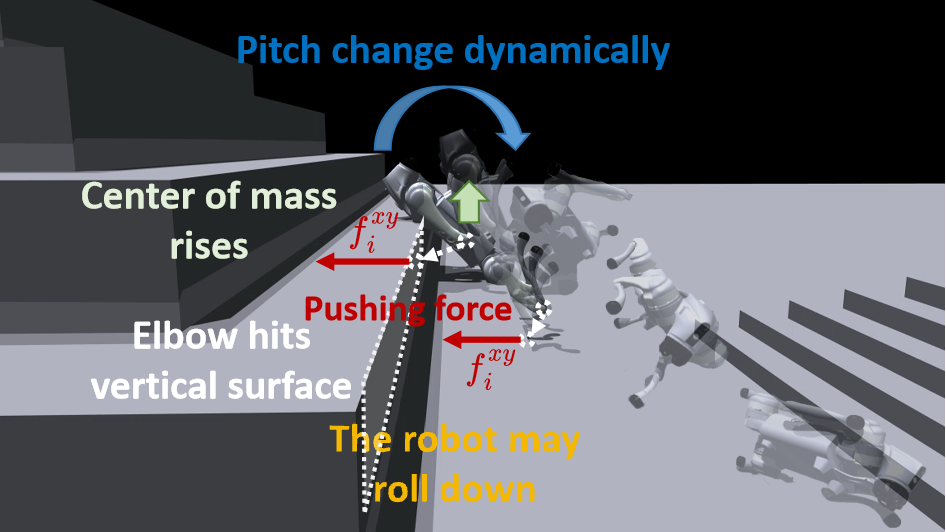}
\caption{\textbf{Unstable vertical contact case:} Elbow contact with vertical surfaces generates high $f_{xy}$ forces, causing significant pitch instability and upward CoM trajectory that can lead to falling from elevated platforms.}
\label{pics:f_xy}
\end{minipage}
\hfill
\begin{minipage}[t]{0.48\columnwidth}
\centering
\includegraphics[width=\textwidth]{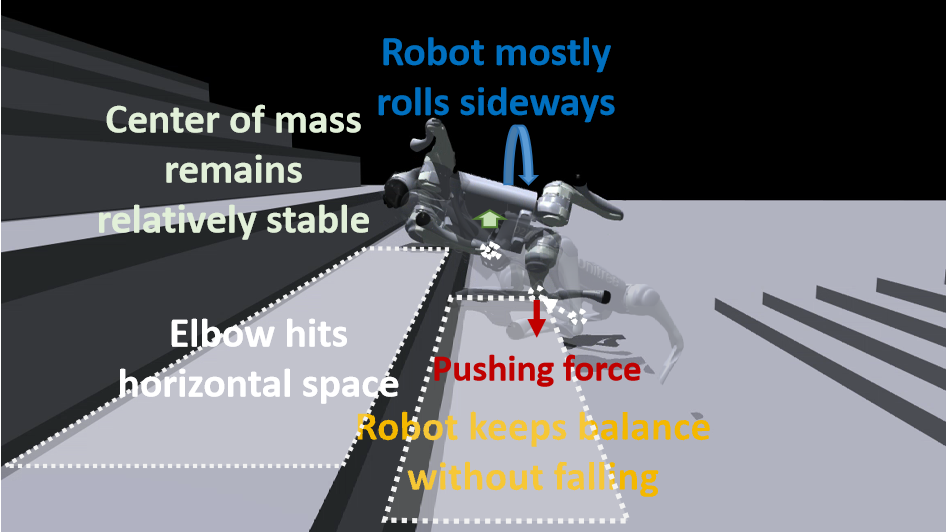}
\caption{\textbf{Stable horizontal contact case:} Elbow contact with horizontal surfaces produces primarily $f_z$ forces, enabling controlled sideways rolling with stable CoM trajectory and minimal positional displacement.}
\label{pics:f_z}
\end{minipage}
\end{figure}

\textcolor{black}{As illustrated in Equation \ref{eq:R_stability}, the $f_{xy}$ penalty component of $R_{\text{stability}}$ directly addresses critical stability challenges during recovery. When the robot's elbow contacts a vertical surface (Fig. \ref{pics:f_xy}), the resulting horizontal force causes the center of mass to rise upward while inducing significant pitch variation. This creates unstable dynamics that can lead to dangerous rolling, particularly on steep terrain. In contrast, contact with horizontal surfaces (Fig. \ref{pics:f_z}) enables the robot to maintain a relatively stable center of mass trajectory while executing controlled sideways rolling motions that facilitate safer recovery.}


\textcolor{black}{By penalizing high horizontal contact forces against vertical surfaces, our stability control reward guides the policy to 1) modulate joint torques when exploring contact points, 2) prefer contact configurations that maintain COM stability, and 3) reduce excessive pitch variations that lead to uncontrolled tumbling to some extent.}

\subsubsection{Motion Constraints}
This component implements standard kinematic constraints on the robot's limb positions, velocities, and accelerations to ensure smooth and stable motion execution. These basic constraints follow conventional robotics practice and help prevent excessive joint movements during recovery.

\begin{table}[htbp]\footnotesize
\caption{Reward terms for quadrupedal recovery}
\centering
\setlength{\tabcolsep}{1pt}
\scriptsize
\begin{tabular}{cccc}
\toprule
Category & Reward & Equation & Weight ($w_i$) \\[0.5ex]
\midrule
\multirow{4}{*}{\makecell[c]{Orientation \\ Posture}}
 & Base Orientation & $ \boldsymbol{\boldsymbol{g}}_{xy}^2$ & -0.5 \\
 & Upright Orientation & $\exp(-\frac{(g_z + 1)^2}{2\epsilon^2})$ & 6.0 \\[0.5ex]
 & Target Posture  & \makecell{$\exp(-(\boldsymbol{q} - \boldsymbol{q}_{stand})^2)$ \\ if $|g_z + 1| < \epsilon$} & 4.0 \\
\midrule
\multirow{2}{*}{\makecell[c]{Contact \\ Management}}
 & Feet Contact & $\sum_{i=1}^4 \mathds{1}_{f_{contact_i}}$ & 0.3 \\[0.5ex]
 & Body Contact & $\mathds{1}_{contact}$ & -0.2 \\[0.5ex]
\midrule
\multirow{2}{*}{\makecell[c]{Stability \\ Control}}
 & Safety Force & $\sum_{i=1}^{4} ||\boldsymbol{f}_{i}^{xy}||_2$ & $-1.0\text{e-}2$ \\[0.5ex]
 & Body-bias & $\text{clip}(||\boldsymbol{p}_{current}^{xy} - \boldsymbol{p}_{init}^{xy}||_2, 0, 4)$ & -0.1 \\[0.5ex]
\midrule
\multirow{6}{*}{\makecell[c]{Motion \\ Constraints}}
 & Position Limits & $\sum_{i=1}^{12} \mathds{1}_{q_i>q_{max}||q_i<q_{min}}$ & -1.0 \\[0.5ex]
 & Angular Velocity Limit & $\max(|\dot{\boldsymbol{q}}| - 0.8, 0)$ & -0.1 \\
 & Joint Acceleration & $\ddot{\boldsymbol{q}}^2$ & $-2.5\text{e-}6$ \\
 & Joint Velocity & $ \dot{\boldsymbol{q}}^2$ & $-1.0\text{e-}2$ \\
 & Action Smoothing & $(\boldsymbol{a}_t - \boldsymbol{a}_{t-1})^2 $ & -0.01 \\
 & Joint Torques & $ \boldsymbol{\tau}^2$ & $-5.0\text{e-}4$ \\
\bottomrule
\end{tabular}
\label{tab:reward_terms}
\end{table}
The Orientation and Posture terms guide upright recovery with significant weights on upright orientation and target posture. Contact Management promotes proper foot placement while penalizing undesired body contacts. Stability Control introduces crucial safety measures by regulating contact forces and limiting positional drift. Motion constraints terms maintain motion feasibility through position limits, velocity constraints, and various smoothing penalties. This comprehensive design enables stable and efficient recovery strategies while ensuring smooth and natural movements across challenging scenarios.

\subsection{Curriculum Learning}
\begin{figure}[h]
    \centering
    \includegraphics[width=0.5\textwidth, height=0.19\textheight]{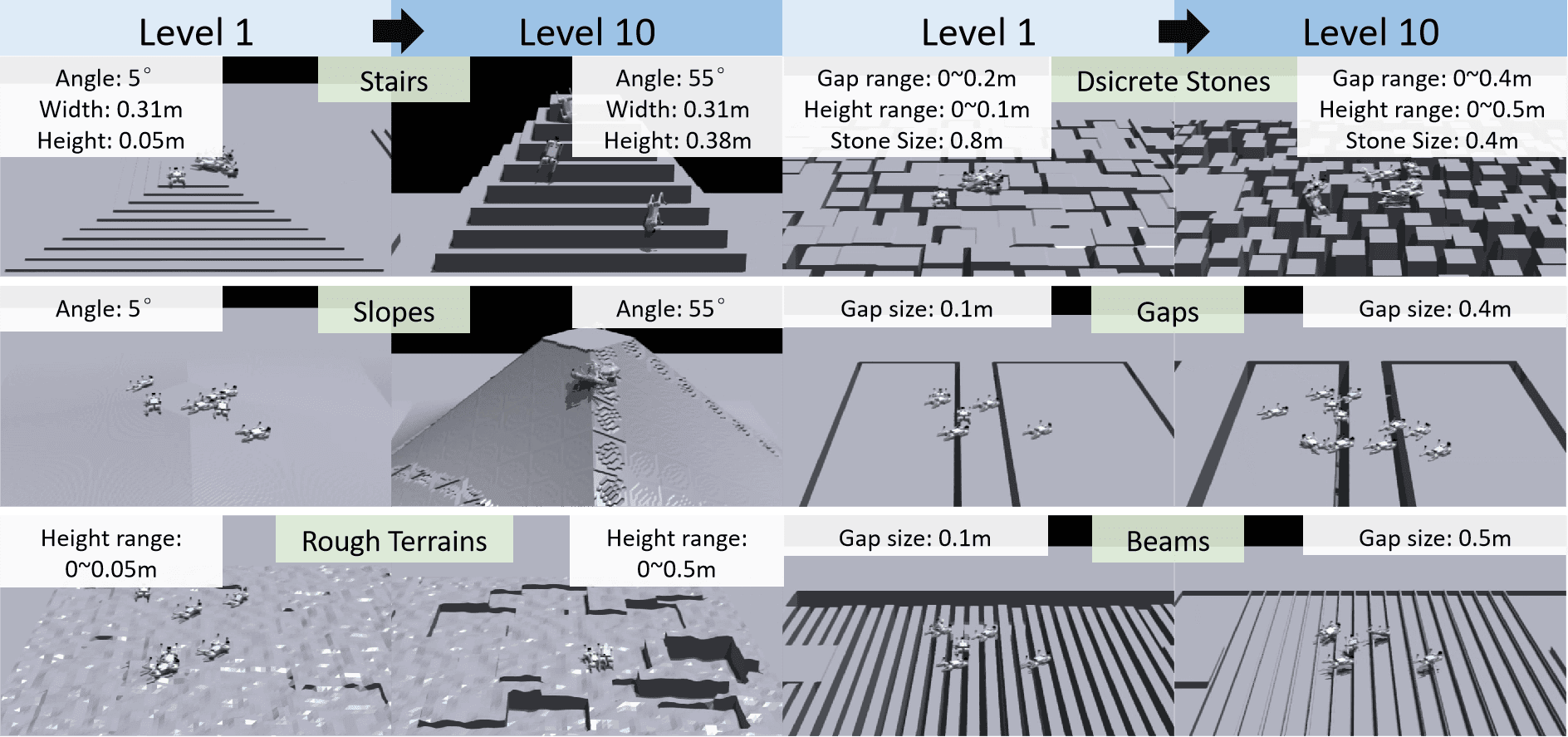}
    \caption{Terrains with 10 different levels.}
    \label{fig:terrains}
\end{figure}

We adopted curriculum learning \cite{wang2021survey} to facilitate progressive learning of recovery abilities across challenging terrains. Each terrain type incorporates randomization in obstacle placement, sizes, and gaps to improve robustness and prevent overfitting. The difficulty of each terrain can be continuously adjusted by modifying parameters such as obstacle density, height variations, and gap sizes, as shown in Fig.~\ref{fig:terrains}.

\begin{figure*}[h]
    \centering
    \includegraphics[width=0.98\textwidth, height=0.18\textwidth]{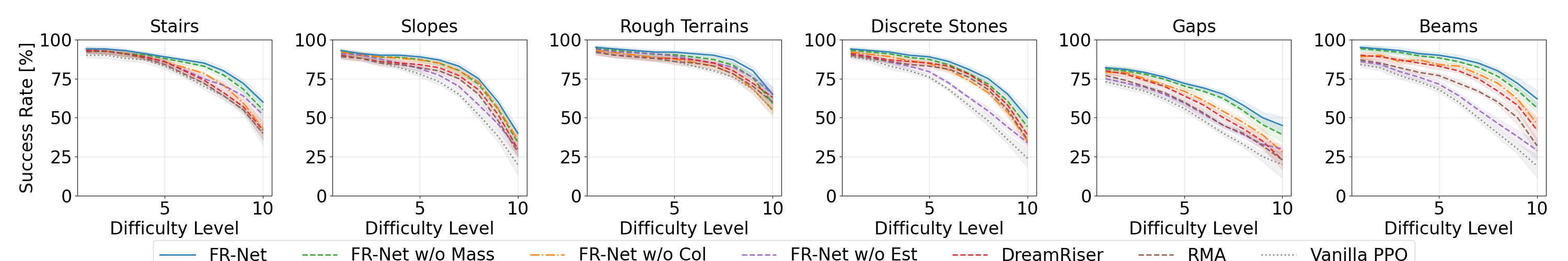}
    \caption{\textcolor{black}{Evaluating the impact of different algorithms on different challenging terrains.}}
    \label{pics:compare_ablation}
\end{figure*}

\section{Experiments}
\subsection{Implementation details}
\subsubsection{Training}Our controller is trained in NVIDIA Isaac Gym \cite{makoviychuk2021isaac}, leveraging its capability for efficient parallel simulations. We train the policy by PPO \cite{schulman2017proximal} with 4,096 parallel domain-randomized morphology quadrupedal robots. Episodes begin with a random supine position and terminate after 350 timesteps. The networks are optimized by the Adam \cite{kingma2014adam}. All the methods above are trained using curriculum learning and reward functions detailed by Fig. \ref{fig:terrains} and table \ref{tab:reward_terms}.

Training is conducted on a workstation with an Intel Core i7-13700K CPU, 32 GB RAM, and an NVIDIA RTX 4070 Ti GPU. To enhance policy robustness, we implement domain randomization as shown in Table \ref{tab:domain_randomization}.

\begin{table}[h]\scriptsize
\fontsize{6.0pt}{7.5pt}\selectfont  
\caption{Domain randomization ranges applied in simulation.}
\label{tab:domain_randomization}
\centering
\renewcommand{\arraystretch}{1.1}
\setlength{\tabcolsep}{2pt}
\begin{tabular}{c|c c c|c c c}
\hline
\textbf{Type} & \textbf{Parameters} & \textbf{Range} & \textbf{Units} & \textbf{Parameters} & \textbf{Range} & \textbf{Units}\\
\hline
\multirow{7}*{Morphology} 
& Trunk Mass & [4.00, 28.00] & kg & Trunk Length & [0.37, 0.65] & m\\
& Trunk Width & [0.09, 0.30] & m & Trunk Height & [0.11, 0.19] & m\\
& Hip Mass & [0.30, 0.69] & kg & Hip Length & [0.03, 0.05] & m\\
& Thigh Mass & [0.60, 4.00] & kg & Thigh Length & [0.21, 0.35] & m\\
& Thigh Width & [0.02, 0.04] & m & Thigh Height & [0.03, 0.05] & m\\
& Calf Mass & [0.10, 0.86] & kg & Calf Length & [0.21, 0.35] & m\\
& Calf Width & [0.016, 0.020] & m & Calf Height & [0.013, 0.019] & m\\
\hline
\multirow{3}*{Control} 
& $K_p$ factor & [20, 80] & N·m/rad & $K_d$ factor & [0.6, 2.0] & N·m·s/rad\\
& Motor strength & [0.9, 1.1] & N·m & COM shift & [-0.05, 0.05] & m\\
& Payload & [-2.0, 2.0] & kg & Motor Friction & [0.2, 1.25] & N·m\\
\hline
\end{tabular}
\end{table}
\subsubsection{Deploying}To demonstrate our method's recovery capacity over challenging terrains and robustness in extremely high-risk obstacles, we implement our method on the Unitree Go2 robot. The control policy is deployed on the robot's onboard NVIDIA Jetson Orin Nano. Our controller operates at 50Hz, receiving joint states and IMU data as input. The deployment process requires minimal adaptation, primarily involving sensor data preprocessing and action scaling to match the hardware specifications.

\subsection{Compared Methods}
We compare our methods with serval baselines and ablations as follows.
\begin{itemize}
    \item \textbf{FR-Net (Ours)}: Training with all modules.
    \item \textbf{FR-Net w/o Mass}: Training without mass distribution estimation.
    \item \textbf{FR-Net w/o Col}: Training without collision estimator.
    \item \textbf{FR-Net w/o Est}: Training without Auto-encoder model.
    \item \textbf{DreamRiser} \cite{nahrendra2023robust}: The policy is trained concurrently with an autoencoder estimating the body velocity and a context vector, without the morphology knowledge.
    \item \textbf{\textcolor{black}{RMA~\cite{kumar2021rma}}}\textcolor{black}{:} \textcolor{black}{An online adaptation approach with a teacher-student framework that enables rapid terrain adaptation through extrinsics prediction.}
    \item \textbf{Vanilla PPO} \cite{schulman2017proximal}: Training with only proprioception.
\end{itemize}
\textcolor{black}{For ablation studies, we fix the random seed to ensure consistent initialization, isolating the impact of different methods. To ensure statistical reliability during final evaluations, we test each method with four random seeds and report the mean success rate alongside the standard deviation.}

\subsection{Analysis of the Success Rate}
\textcolor{black}{We assess policy robustness across ten difficulty levels, with increasing complexity in environmental parameters. Each level deploys 2048 robots with diverse morphologies (e.g., Go2, Lite3, Spot, Aliengo, and Go1). Success is defined as achieving a stable upright pose within five seconds, with position displacement under 1 meter. While all methods perform well at lower difficulty levels, performance gaps widen significantly as complexity increases.}

\textcolor{black}{Figure~\ref{pics:compare_ablation} illustrates FR-Net's superior performance across varying terrains. While all algorithms achieve high success rates ($>90\%$) at low difficulty levels (1--3), performance differences become evident as difficulty increases. At maximum difficulty, FR-Net attains the highest success rates on Rough Terrains (69.5\%) and Beams (64.1\%), surpassing Vanilla PPO by 21.8\% and 25.3\%, respectively. On Stairs and Slopes, FR-Net achieves $\geq 58\%$ success at level 10, while alternatives fall below $45\%$. In more demanding terrains (Discrete Stones, Gaps, and Beams), FR-Net outperforms DreamRiser by 8--12\% and Vanilla PPO by 13--25\%. Ablation studies further show FR-Net's robustness: removing state estimation reduces Beam performance by 12.7\%, and omitting mass consideration lowers Gaps success from 46.8\% to 35.5\%. These results confirm FR-Net's ability to navigate complex environments effectively.}

\subsection{Effects of Stability Reward}

\begin{figure}[h]
   \centering
   \includegraphics[width=0.49\textwidth, trim=0 0 0 0,clip]{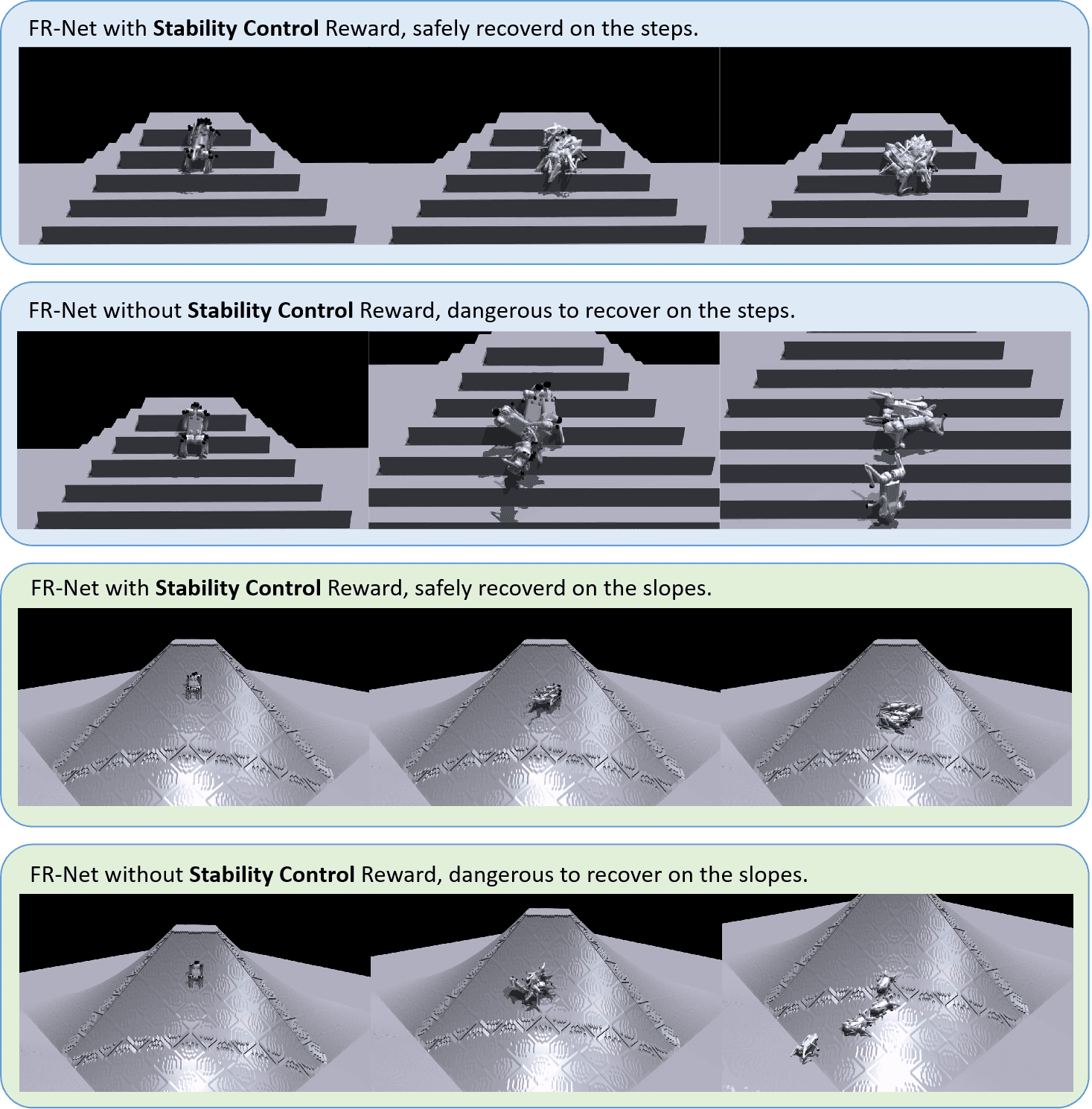}
   \caption{Our methods demonstrate exceptional performance on tilted
 stairs and slopes. All dogs are positioned in a supine posture on a 40° incline.}
   \label{fig:rec_risky}
\end{figure}

We further investigate the effectiveness of our proposed reward functions with stability control reward. As shown in Fig.~\ref{fig:rec_risky}, the FR-Net with these rewards demonstrates remarkable performance on challenging terrains. On 40° stairs, the robot maintains lateral displacement within 0.8 m, while on 40° slopes, it limits displacement to 1.2 m. This controlled recovery behavior effectively prevents dangerous tumbling on steep terrains, significantly enhancing safety during recovery maneuvers. The comparison clearly illustrates that without these rewards, the robot exhibits unstable behaviors and dangerous rolling motions, particularly evident in the bottom sequences of both stairs and slope scenarios.

\subsection{Real-World Experiments} 
We extensively validate our approach through real-world experiments using the Unitree Go2 quadrupedal robot across eleven diverse challenging scenarios, as shown in Fig.~\ref{fig:real_results}. The testing environments range from natural terrains like grass fields and small stones to structured obstacles including stairs, air beams, and hazardous gaps. Our method demonstrates robust recovery capabilities across all scenarios, successfully bridging the reality gap from simulation to real-world deployment. \textcolor{black}{To mitigate aggressive motions learned through RL, scaling actions by a decay factor and adjusting PD parameters (higher derivative gain $k_d$, lower proportional gain $k_p$) may help generate smooth motions during deployment.}

\begin{figure}[h]
    \centering
    \includegraphics[width=0.48\textwidth]{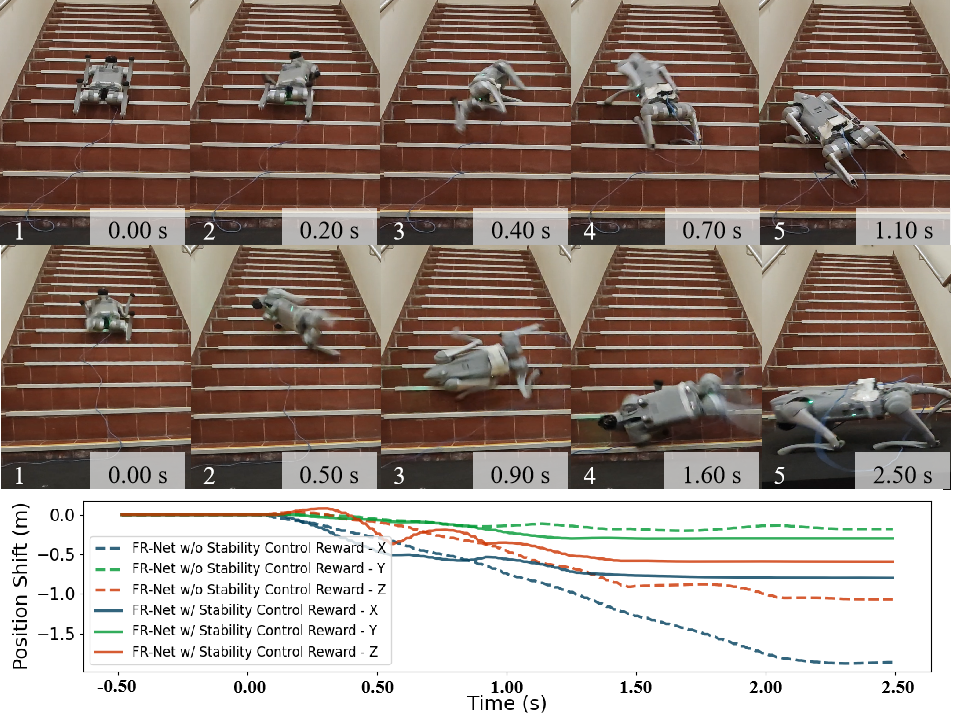}
    \caption{\textcolor{black}{Comparative recovery sequences on 38.5° stairs with and without stability control reward. Bottom: corresponding position shift timeline.}}
    \label{fig:rec_stairs}
\end{figure}

Particularly noteworthy performances are observed in three highly challenging scenarios. Fig.~\ref{fig:rec_stairs} presents a detailed comparison of recovery behaviors on stairs. The top sequence shows our method with the Stability Control reward, demonstrating controlled recovery that prevents downward tumbling. In contrast, the bottom sequence shows the baseline method without this reward, resulting in dangerous rolling motions down the stairs. This comparison clearly validates the effectiveness of our reward design in maintaining safer recovery behaviors on hazardous terrains.
\begin{figure}[h]
    \centering
    \includegraphics[width=0.48\textwidth]{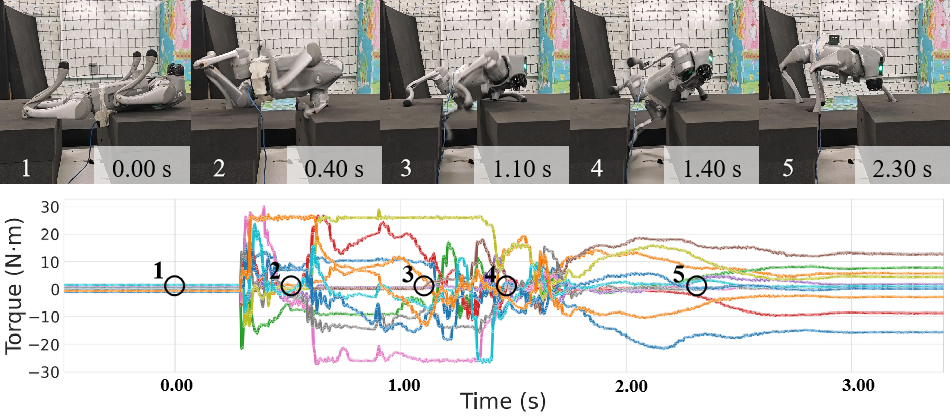}
    \caption{\textcolor{black}{Recovery sequence on challenging gaps (t = 0.00 s to 2.30 s) with corresponding joint torque profiles.}}
    \label{fig:rec_gaps}
\end{figure}

\textcolor{black}{We further evaluate our method's robustness on challenging gaps, as shown in Fig.~\ref{fig:rec_gaps}. The sequential snapshots demonstrate the robot's recovery capability when encountering hazardous situations. At t = 1.40 s, two legs step into the gap. While our method cannot precisely predict contact geometry in these complex scenarios with hidden surfaces, the robot still manages to execute effective recovery motions (t = 1.40 s to 2.30 s) based on the available proprioceptive feedback. Through a combination of contact detection and reactive adaptation, the robot adjusts its limb positioning to leverage the detected support surfaces.}

\begin{figure}[h]
    \centering
    \includegraphics[width=0.48\textwidth]{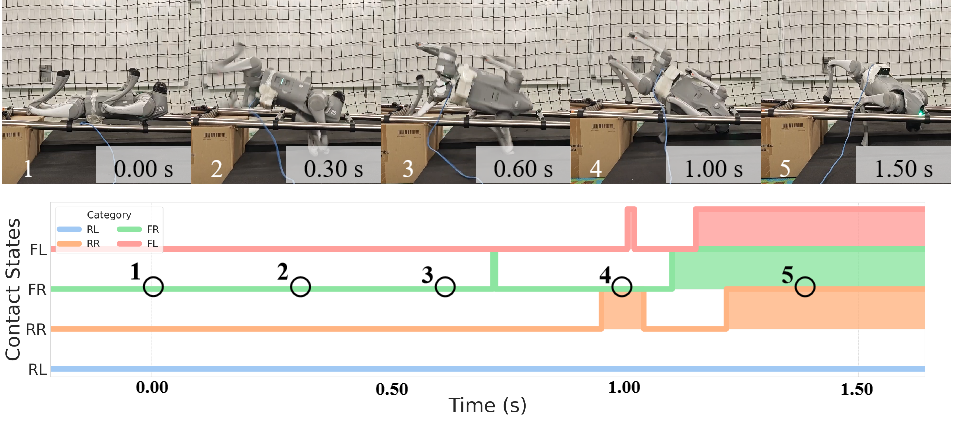}
    \caption{\textcolor{black}{Recovery sequence on suspended beams (t = 0.00 s to 1.50 s) with corresponding foot contact states.}}
    \label{fig:rec_beam}
\end{figure}

\textcolor{black}{We evaluate our method on suspended beams, as shown in Fig.~\ref{fig:rec_beam}. From an initial unstable position (t = 0.00 s), the robot executes a controlled recovery despite the challenging confined environment. The contact visualization reveals that only three legs engage with the ground, with the rear left foot intentionally maintained off-ground. During the critical phase (t = 0.40-1.00 s), our approach coordinates precise weight redistribution among supporting legs, particularly using the front right leg for stabilization. This strategic planning prevents mechanical wedging between beams and achieves a stable final pose (t = 1.50 s). The successful recovery with partial support demonstrates our method's capabilities in confined scenarios.}

\begin{figure}[h]
    \centering
    \includegraphics[width=0.48\textwidth]{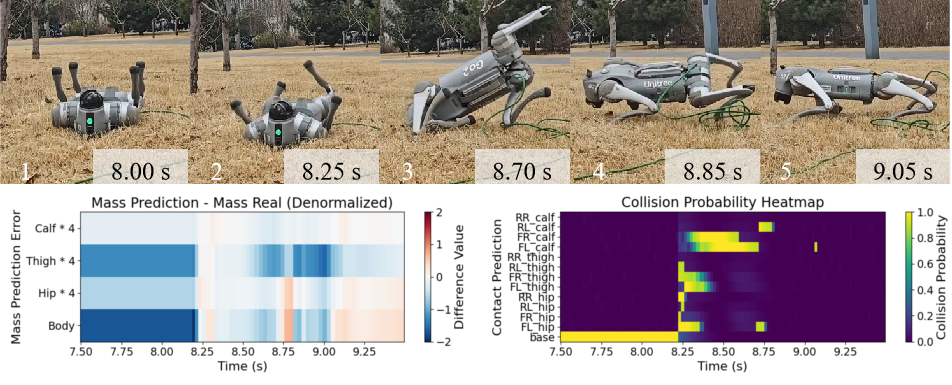}
    \caption{\textcolor{black}{Analysis of MCP output. \textbf{Top}: Sequential snapshots showing recovery transition from 8.00 s to 9.05 s. \textbf{Bottom Left}: Heat map of mass prediction error between estimated and actual values across body components. \textbf{Bottom Right}: Contact probability heat map showing predicted collision likelihood for different leg segments during recovery.}}
    \label{fig:mcp_output}
\end{figure}

\textcolor{black}{Fig.~\ref{fig:mcp_output} demonstrates our Mass-Contact Predictor's performance during robot recovery. In the static phase (before 8.25 s), mass estimation accuracy is limited due to minimal joint forces, visible as blue regions in the bottom left heat map. During dynamic motion (8.25-9.05 s), prediction accuracy improves significantly with increased joint loads. Note that *4 represents the sum of four joints. The normalized scale is 15 for the body and 5 for the limbs. The bottom right visualization shows contact probability predictions, demonstrating how our approach anticipates physical interactions during recovery without external sensing.}

\subsection{\textcolor{black}{Validation on Humanoid Robot}}
\begin{figure}[h]
    \centering
    \includegraphics[width=0.48\textwidth, height=0.19\textwidth]{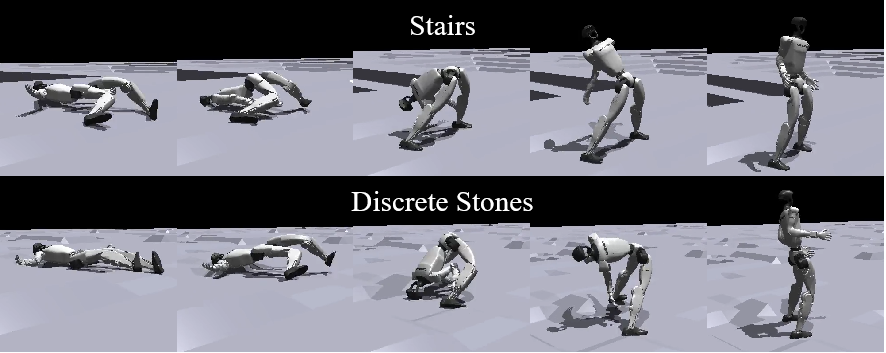}
    \caption{\textcolor{black}{Recovery motions of the humanoid robot Unitree G1 in simulation.}}
    \label{fig:human1}
\end{figure}

\textcolor{black}{We test our framework on the humanoid robot Unitree G1 in simulation, as shown in Fig.~\ref{fig:human1}, using our proposed reward functions together with the ones that are specific for humanoid robots \cite{he2025learning,huang2025learning}. Detailed reward designs can be found in the supplementary material available at~\cite{appendix2025}. The robot successfully recovers on irregular terrains, demonstrating the adaptability of our methods.}


\section{\textcolor{black}{CONCLUSION AND LIMITATIONS}}
\textcolor{black}{We proposed FR-Net, a robust framework for quadrupedal fall recovery in challenging terrains. By integrating a Mass-Contact Predictor and carefully designing reward functions, FR-Net achieved safe, efficient, and generalizable recovery motions. Extensive experiments demonstrated its superior performance in complex environments, with effective real-world deployment on the Go2 robot. This work demonstrates the potential of learning-based approaches for enhancing legged robot autonomy and adaptability.}

\textcolor{black}{Despite these achievements, our approach currently estimates only body and joint collisions rather than precise contact locations on the robot's surface, particularly in scenarios like gaps where the system detects general collision states but not exact terrain geometry. Integration of additional tactile sensors could address this limitation in future work.}

\bibliographystyle{IEEEtran}
\bibliography{reference}

\end{document}